\documentclass[10pt,twocolumn,letterpaper]{article}
\usepackage{amsthm,amsmath,amssymb}
\usepackage{mathrsfs}
\usepackage{iccv}
\usepackage{times}
\usepackage{epsfig}
\usepackage{graphicx}
\usepackage{amsmath}
\usepackage{amssymb}
\usepackage{multirow}
\usepackage{caption}
\usepackage{overpic}
\usepackage{bbding}
\usepackage{color}
\usepackage[breaklinks=true,bookmarks=false]{hyperref}
\iccvfinalcopy % *** Uncomment this line for the final submission

\def\iccvPaperID{2714} % *** Enter the ICCV Paper ID here

\newcommand{\tabincell}[2]{\begin{tabular}{@{}#1@{}}#2\end{tabular}}

\ificcvfinal\fi

\begin{document}

%%%%%%%%% TITLE
% \title{Benchmarking Video Instance Lane Detection:\\
% Annotated Dataset and A Multi-level Memory Aggregation Network}

\title{VIL-100: A New Dataset and A Baseline Model for \\ Video Instance Lane Detection}

\author{Yujun Zhang$^{1}$\footnotemark[1], Lei Zhu$^{2}$\footnotemark[1], Wei Feng$^{1}$\footnotemark[2], Huazhu Fu$^{3}$, Mingqian Wang$^{1}$,  \\ Qingxia Li$^{4}$, Cheng Li$^{1}$, and Song Wang$^{1,5}$ \\
$^1$ Tianjin University, 
$^2$ University of Cambridge, 
$^3$ Inception Institute of Artificial Intelligence, \\
$^4$ Automotive Data of China (Tianjin) Co., Ltd, 
$^5$ University of South Carolina
}

%\author{Yujun Zhang$^{1, \ast}$, Lei Zhu$^{2}$\thanks{Yujun Zhang and Lei Zhu are the joint first authors of this work.}, Wei Feng$^{1}$\thanks{Corresponding author (wfeng@ieee.org)}, Huazhu Fu$^{3}$, Mingqian Wang$^{1}$,  \\ Qingxia Li$^{4}$, Cheng Li$^{1}$, and Song Wang$^{1,5}$ \\
%$^1$ Tianjin University, 
%$^2$ University of Cambridge, 
%$^3$ Inception Institute of Artificial Intelligence, \\
%$^4$ Automotive Data of China (Tianjin) Co., Ltd, 
%$^5$ University of South Carolina
%}

\twocolumn[{
  \maketitle
  \begin{center}
    \vspace{-23pt}
    \begin{overpic}[width=1.0\linewidth]{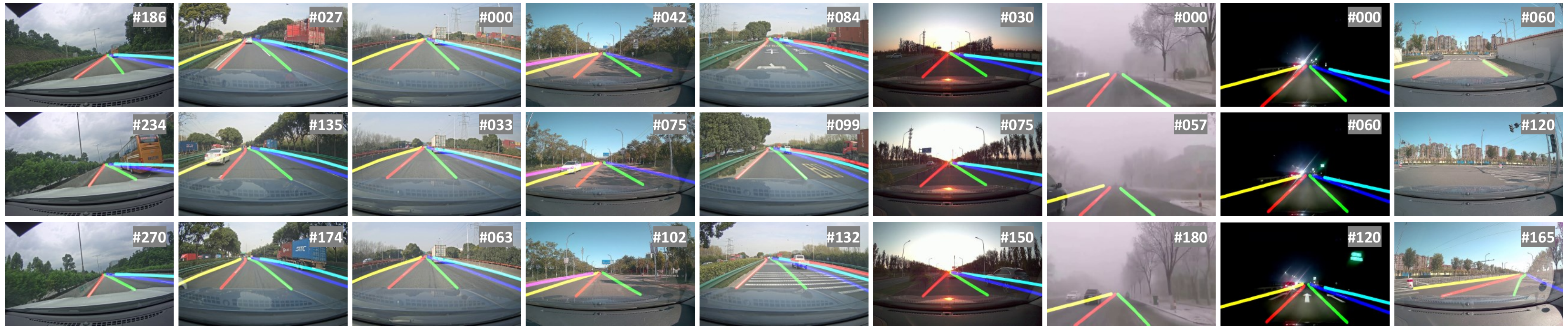}
    \end{overpic}
    \vskip -5pt
    \captionof{figure}{
    Our proposed Video Instance Lane Detection (VIL-100) dataset contains different real traffic scenarios, and provides the high-quality instance-level lane annotations. 
    }
    \label{fig:dataset}
    \vspace{-3mm}
  \end{center}
}]

\maketitle
% Remove page # from the first page of camera-ready.
\ificcvfinal\thispagestyle{empty}\fi

\renewcommand{\thefootnote}{\fnsymbol{footnote}} \footnotetext[1]{Yujun Zhang and Lei Zhu are the joint first authors of this work.}
\footnotetext[2]{Wei Feng (wfeng@ieee.org) is the corresponding author.}

\begin{abstract}
 Lane detection plays a key role in autonomous driving. While car cameras always take streaming videos on the way, current lane detection works mainly focus on individual images (frames) by ignoring dynamics along the video.  
 In this work, we collect a new video instance lane detection (VIL-100) dataset, which contains 100 videos with in total 10,000 frames, acquired from different real traffic scenarios. 
 All the frames in each video are manually annotated to a high-quality instance-level lane annotation, and a set of frame-level and video-level metrics are included for quantitative performance evaluation. 
 Moreover, we propose a new baseline model, named multi-level memory aggregation network (MMA-Net), for video instance lane detection. 
 In our approach, the representation of current frame is enhanced by attentively aggregating both local and global memory features from other frames. 
 Experiments on the new collected dataset show that the proposed MMA-Net outperforms state-of-the-art lane detection methods and video object segmentation methods.
 We release our dataset and code at \url{https://github.com/yujun0-0/MMA-Net}.

\end{abstract}

\section{Introduction}
\label{sec::introduction}

%%%%%%%%%%%%%%%%%%%%%%%%%%%%%%%%%%%%%%%%%%%%%%%
% Background
%%%%%%%%%%%%%%%%%%%%%%%%%%%%%%%%%%%%%%%%%%%%%%%
In recent years, autonomous driving has received numerous attention in both academy and industry. 
One of the most fundamental and challenging task  is lane detection in traffic scene understanding.
%Instance lane detection aims to automatically identify all lane regions, and assign different labels for each lane.
Lane detection assists car driving and could be used in advanced driving assistance system (ADAS)~\cite{1603550,r5,TANG2021107623}.
%, such as lane departure, trajectory planning decision, and son on.
% Lane detection is utilized to guide vehicles and could be used in 
However, accurately detecting lanes in real traffic scenarios is very challenging, due
to many harsh scenarios, e.g., severe occlusion, bad weather
conditions, dim or dazzle light.

%%%%%%%%%%%%%%%%%%%%%%%%%%%%%%%%%%%%%%%%%%%%%%%
% Related work
%%%%%%%%%%%%%%%%%%%%%%%%%%%%%%%%%%%%%%%%%%%%%%%
With the advancement of deep learning, lane detection has achieved significant progress in recent years by annotating and training on large-scale real data~\cite{r17,r14,r11,LaneNet2018,SCNN2018}.
However, most of the existing methods are focused on image-based lane detection, while in autonomous driving, car camera always collects streaming videos. It is highly desirable to extend deep-learning  based lane detection from image level to video level since the latter can leverage temporal consistency to resolve many in-frame ambiguities, such as occlusion, lane damage etc. The major obstacle for this extension is the lack of a video dataset with appropriate annotations, both of which are essential for deep network training. 
Existing lane datasets (e.g., TuSimple ~\cite{Tusimple2017}, Culane~\cite{SCNN2018}, ApolloScape~\cite{ApolloScape2019} and BDD100K~\cite{BDD100K2020}) either support only image-level
detection or lack temporal instance-level labels.
%%
%Most existing video lane detection datasets only cover relatively simple scenarios and/or are insufficiently annotated. For example, none of existing video dataset, e.g., TuSimple Benchmark~\cite{Tusimple2017}, Culane~\cite{SCNN2018}, ApolloScape~\cite{ApolloScape2019} and BDD100K~\cite{BDD100K2020}, provides the important instance-level lane annotations all across the video. 
%%
However, according
to the United Nations Regulation No.157~\cite{UN157} for autonomous
and connected vehicles, continuous instance-level
lane detection in videos is indispensable for regular/emergency
lane change, trajectory planning, autonomous navigation, etc.

% What's more, deep-learning methods have not been much exploited for video lane detection due to the lack of such a dataset.

%A number of image lane detection methods have been developed in the past decades. 
%%
%With the development of deep learning techniques, Convolutional Neural Networks (CNNs) have been widely used and  achieved a gratifying progress in lane detection by learning discriminate deep features from annotated data~\cite{r17,r14,r11,LaneNet2018,SCNN2018}.
%%
%However, in striking contrast with the flourishing development of image lane detection, much fewer works have
%been explored in lane detection over dynamic traffic scenes.
%On the other hand, we also notice that video processing has become an urgent topic in recent years, and a lot of methods were proposed for video object detection and segmentation~\cite{10.1145/3391743,Liu2020,9345705}. 
%What makes video lane detection lag far behind these video processing tasks? 
%Compared with lane detection of a single image,
%video lane detection (VLD) needs to utilize temporal information to identify lane instances of each video frame.
%Although there exist multiple datasets for image lane detection, video object detection and segmentation, such \textit{standard widespread benchmark (with a sufficient number of video clips, covering diverse content with instance-level lane annotation) is still missing for video instance lane detection.}
%What’s more, CNN-based methods have not been exploited for this problem due to the lack of such a dataset.

%%%%%%%%%%%%%%%%%%%%%%%%%%%%%%%%%%%%%%%%%%%%%%%
To address above issues, in this work, \textbf{we first collect a new video instance lane detection (VIL-100) dataset} (see Figure~\ref{fig:dataset} for examples). 
It contains 100 videos with 10,000 frames, covering %10 kinds of lines
10 common line-types, multiple lane instances, various driving scenarios, different weather and lighting conditions.
All the video frames are carefully annotated with high-quality instance-level lane masks, which could facilitate the community to explore further in this field.  
% . To the best of our knowledge, this is the first learning-oriented dataset for video lane detection, which could facilitate the community to explore further in this field.  
%%%%%%%%%%%%%%%%%%%%%%%%%%%%%%%%%%%%%%%%%%%%%%%
Second, \textbf{we develop a new baseline model, named multi-level memory aggregation network (MMA-Net).}
Our MMA-Net leverages both local memory and global memory information to enhance multiple CNN features extracted from the current key frame. 
To be specific, we take past frames of the original video to form a local memory and past frames of a shuffled ordered video as a global memory, and the current video frame as the query is segmented using features extracted from video frames of both local memory and global memory.
A local and global memory aggregation (LGMA) module is devised to attentively aggregate these multi-level memory features, and then all CNN features are integrated together to  produce the video instance lane detection results.
Finally, \textbf{we present a comprehensive evaluation of 10 state-of-the-art models on our VIL-100 dataset}, making it the most complete video instance-level lane detection benchmark. 
Results show that our MMA-Net significantly outperforms existing methods, including single-image lane detectors~\cite{r17,r14,r11,LaneNet2018,SCNN2018}, and video instance object segmentation methods~\cite{GAM,TVOS,o6,AFB,RVOS}. 
%%
% In summary, our work provides a new benchmark for video instance lane detection, and a developing novel memory aggregation network.

%Our main contributions of this work are three-folded: (1) We propose an adaptive and efficient feature aggregation module to maintain most useful temporal information for video lane segmentation. (2) We collect and annotate the first new video instance lane detection (VIL) dataset, which could further contribute the community to explore more in this field. (3) We demonstrate the effectiveness of our method on video lane detection, which is the first work to exploit the temporal consistency.
\section{Related Works}
\label{sec:relatedworks}

\noindent \textbf{Lane detection datasets.}
Large-scale datasets are important for deep learning methods. Several public datasets, such as Caltech-Lanes~\cite{Calteh2008}, TuSimple~\cite{Tusimple2017}, Culane~\cite{SCNN2018}, ApolloScape~\cite{ApolloScape2019} and BDD100K~\cite{BDD100K2020}, have been used for lane detection study.
Table~\ref{table:datasets_com} provides a comparison of VIL-100 and other public available datasets from different perspectives.
Caltech-Lanes only contains 1,224 images and is usually not used for  training deep networks, while  TuSimple and Culane provide large-scale image data with instance-level lane annotations. However, we are interested in video instance lane detection in this paper, for which both TuSimple and Culane are not applicable. BDD100K and ApolloScape are two large-scale video datasets for driving. However, these two datasets do not provide annotations of lane instances  -- on each frame, multiple lanes of the same kind are not separated and annotated with one label. 
Lane-instance detection is important for regular/emergency
lane change, trajectory planning, autonomous navigation in autonomous driving, and we provide video-level lane-instance annotations on our collected VIL-100 dataset. 
% to-do: mention "accurate lane instance detection is critical for navigation in autonomous driving." in introduction.   
%For most of these existing datasets, there are no more than four lanes annotated at a time. 
In our VIL-100, we increase six lanes annotated at a time by including more complex scenes. In addition, we annotate the relative location of each lane to the camera-mounted vehicle in VIL-100 and such location information was not annotated on any previous datasets.

% Recently, the large scale TuSimple and Culane consists of 6,408 and 133,235 images with annotations respectively, which are widely used for deep learning based image lane detection methods.  Both of them provide instance-level lane annotation, and the max lane numbers is 4. TuSimple contains a single scene (sunny day), whereas Culane includes challenging scenes and labels the occluded parts. BDD100K and ApolloScape are large-scale and diverse datasets of public driving, which with video-level annotations, but there is no marked lane instance. As for our VIL-100, not only does it contain rich scenes and lanes, but the maximum number of lanes has been increased to 6. In addition, the lane location is tagged, which can be useful in vehicles lane changing and other cases.

\noindent \textbf{Lane detection.}
Early lane detection methods mostly relied on  hand-crafted features, such as color~\cite{r1,r2}, edge~\cite{r3,r4,r5} and texture~\cite{r6}. Recently, the use of deep neural networks~\cite{r19,r7,LaneNet2018} has significantly boosted the lane detection performance.  %In~\cite{r7}, a CNN initially trained for vehicle detection is extended to lane detection by including new categories of lanes. 
%To improve robustness to various challenging scenarios, 
In VPGNet~\cite{r8}, vanishing points were employed to guide a multi-task network training for lane detection. 
%In~\cite{LaneNet2018}, the binary lane segmentation is further clustered by using Mean-Shift algorithm to reach lane-instance segmentation. 
SCNN~\cite{SCNN2018} specifically considers the thin-and-long shape of lanes by passing message between adjacent rows and columns at a feature layer. 
SAD~\cite{r11} and inter-region affinity KD~\cite{r12} further adopt the knowledge distillation to improve lane detection.
PolyLaneNet~\cite{r13} formulates the instance-level lane detection as a polynomial-regression problem, and UFSA~\cite{r14} provides ultra-fast lane detection by dividing the image into grids and then scanning grids for lane locations. Recently, GAN~\cite{r16} and Transformer~\cite{r17} are also used for detecting lanes. 
Different from the above methods that detect lanes from individual images, this paper addresses video lane detection, for which we propose a new VIL-100 dataset and a baseline MMA-NET method.

\noindent \textbf{Video object segmentation.}
General-purpose video object segmentation (VOS) methods can also be adapted for video lane detection. 
Existing VOS methods can be divided into two categories: zero-shot methods and one-shot methods. They differ in that the latter requires the true segmentation on the first frame while the former does not. 
For zero-shot VOS, traditional methods are usually based on heuristic cues of motion patterns~\cite{z1,z2}, proposals~\cite{z3,z4} and saliency~\cite{z5,z6}. Recent deep-learning based methods include the two-stream FCN~\cite{z7,z8} that integrates
the target appearance and motion features. Recurrent networks~\cite{z9, RVOS} are also used for video segmentation by considering both spatial and temporal consistency. 
For the one-shot VOS, earlier methods usually compute classical optical flow~\cite{o1,o2,o3} for inter-frame label propagation. Recent deep-network based methods~\cite{o4,o5,o6,GAM,TVOS} include GAM~\cite{GAM}, which integrates a generative model of the foreground and background appearance to avoid online fine-tuning. TVOS~\cite{TVOS} suggests a deep-learning based approach for inter-frame label propagation by combining the historical frames and annotation of the first frame.
STM~\cite{o6} uses the memory network to adaptively select multiple historical frames for helping the segmentation on the current frame. STM exhibits superior performance on many available tasks and we take it as the baseline to develop our proposed MMA network. 

\section{Our Dataset}
\label{section3:dataset}

% \begin{figure*}
% 	\begin{center}
% 		\includegraphics[width=1\linewidth]{figs/dataset.png}
% 	\end{center}
% 	\caption{The examples of the proposed Video Instance Lane Detection (VIL) dataset, with pixel-level lane annotations.}
% 	\label{fig:dataset}
% \end{figure*}

% Please add the following required packages to your document preamble:
% \usepackage{multirow}

\begin{table*}[]
	\caption{Comparisons of our dataset and existing lane detection datasets. `$\#$Frames' column shows the number of annotated-frames and the total number of frames. While TuSimple provides a video dataset, it only annotates the last frame of each video and supports image-level lane detection.}
	\vspace{-5mm}
	\begin{center}
		\resizebox{0.95\textwidth}{!}{
			\begin{tabular}{c|c|ccc|cccccc}
				\hline
				\multirow{2}{*}{Dataset}& 
				\multirow{2}{*}{Lane detecion on}& 
				\multicolumn{3}{c|}{Size}& 
				\multicolumn{5}{c}{Diversity}\\ \cline{3-11} %
				& & \begin{tabular}[c]{@{}c@{}}$\#$Videos\end{tabular} & \begin{tabular}[c]{@{}c@{}}$\#$Frames\end{tabular}  & \begin{tabular}[c]{@{}c@{}}Average \\ Length\end{tabular}  & \begin{tabular}[c]{@{}c@{}} Instance-level \\ Annotation\end{tabular}& \begin{tabular}[c]{@{}c@{}}Maximum  \\ $\#$Lanes \end{tabular}  &
				\begin{tabular}[c]{@{}c@{}}Lane\\ Location\end{tabular}
				
				& Line-type 		& Scenario      & Resolution                      \\ \hline
				\begin{tabular}[c]{@{}c@{}}Caltech Lanes\\ 2008~\cite{Calteh2008}\end{tabular} & Video & 4  & \begin{tabular}[c]{@{}c@{}}1224/\\ 1224\end{tabular} & -   & \Checkmark & 4  & -  & -   & \begin{tabular}[c]{@{}c@{}}light traffic,\\ day\end{tabular}    & 640 $\times$ 480 \\ \hline
				\begin{tabular}[c]{@{}c@{}}TuSimple\\ 2017~\cite{Tusimple2017}\end{tabular} & Image     & 6.4K  & \begin{tabular}[c]{@{}c@{}}6.4K/\\ 128K\end{tabular} & 1s       & \Checkmark & 5 & - & - & \begin{tabular}[c]{@{}c@{}}light traffic,\\ day\end{tabular}& 1280 $\times$ 720  \\ \hline
				\begin{tabular}[c]{@{}c@{}}Culane\\ 2017~\cite{SCNN2018}\end{tabular}&Image & - & \begin{tabular}[c]{@{}c@{}}133K/\\ 133K\end{tabular} & -   &\Checkmark &4 &-   & -& \begin{tabular}[c]{@{}c@{}}multi-weather,\\ multi-traffic scene,\\ day \& night\end{tabular} & 1640 $\times$ 590  \\ \hline
				\begin{tabular}[c]{@{}c@{}}ApolloScape\\ 2019~\cite{ApolloScape2019}\end{tabular} &Video  & 235   & \begin{tabular}[c]{@{}c@{}}115K/\\ 115k\end{tabular} & 16s   &\XSolidBrush &-  &- &13 & \begin{tabular}[c]{@{}c@{}}multi-weather,\\ multi-traffic scene,\\ day \& night\end{tabular} & 3384 $\times$ 2710  \\ \hline
				\begin{tabular}[c]{@{}c@{}}BDD100K\\ 2020~\cite{BDD100K2020}\end{tabular}  &Video  & 100K  & \begin{tabular}[c]{@{}c@{}}100K/\\ 120M\end{tabular} & 40s & \XSolidBrush  &  -  & - & 11 & \begin{tabular}[c]{@{}c@{}}multi-weather,\\ multi-traffic scene,\\ day \& night\end{tabular} & 1280 $\times$ 720  \\ \hline \hline
				\begin{tabular}[c]{@{}c@{}}VIL-100(ours) \\2021\end{tabular}  &Video & 100   & \begin{tabular}[c]{@{}c@{}}10K/\\ 10K\end{tabular}   & 10s & \Checkmark & 6 & 8    & 10 & \begin{tabular}[c]{@{}c@{}}multi-weather,\\ multi-traffic scene,\\ day \& night\end{tabular} & \begin{tabular}[c]{@{}c@{}c@{}}640 $\times$ 368 \\ $\sim$\\1920 $\times$ 1080\end{tabular} \\ \hline

			\end{tabular}	
		}
	\end{center}
	\vspace{-3mm}
	\label{table:datasets_com}
\end{table*}

\begin{figure}
	\begin{center}
		\includegraphics[width=1\linewidth]{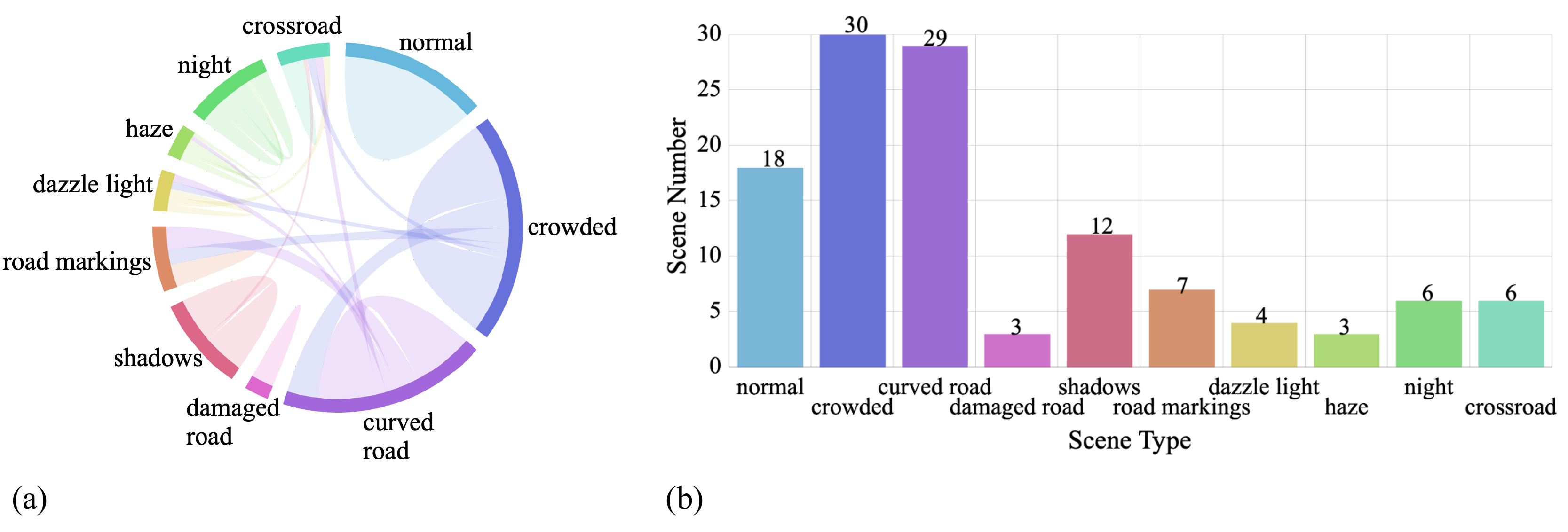}
	\end{center}
	\vspace{-5mm}
	\caption{(a) Co-occurrence of different scenarios. (b) Scenario statistics of the proposed VIL-100.}
	\vspace{-3mm}
	\label{fig:scenarios}
\end{figure}

\begin{figure}
	\begin{center}
		\includegraphics[width=1\linewidth]{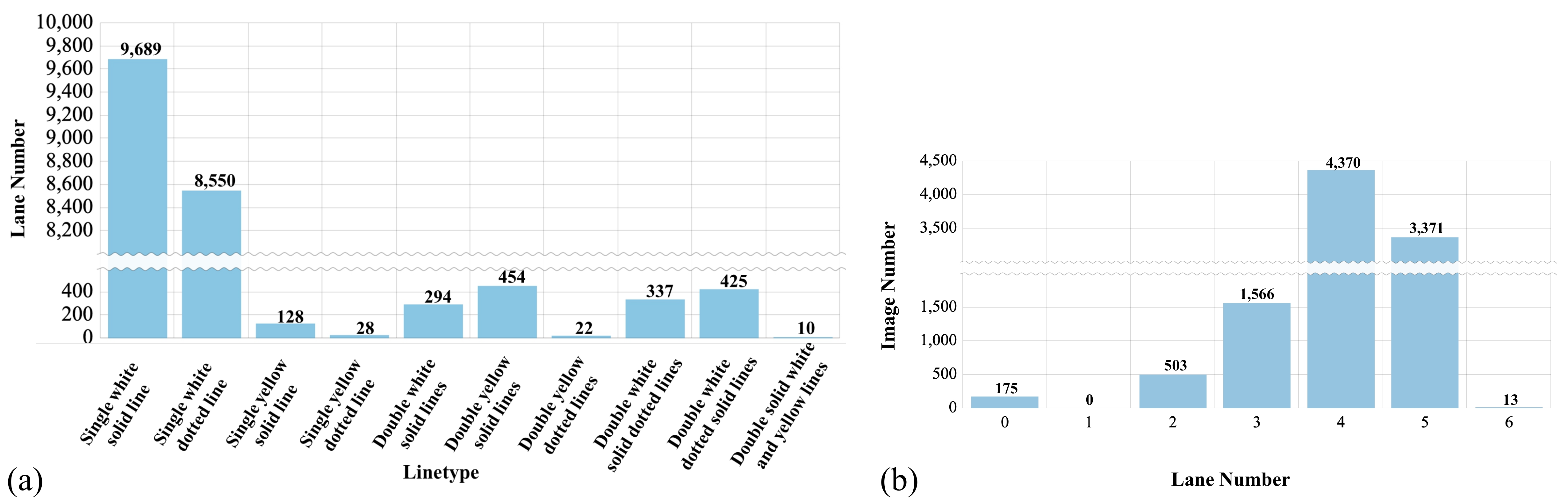}
	\end{center}
	\vspace{-5mm}
	\caption{(a) Distributions of 10 line-types in our VIL-100. (b) Video frame statistics of the number of annotated lanes in our VIL-100.}
	\vspace{-5mm}
	\label{fig:lanes}
\end{figure}

\begin{figure*}
	\begin{center}
		\includegraphics[width=0.85\linewidth]{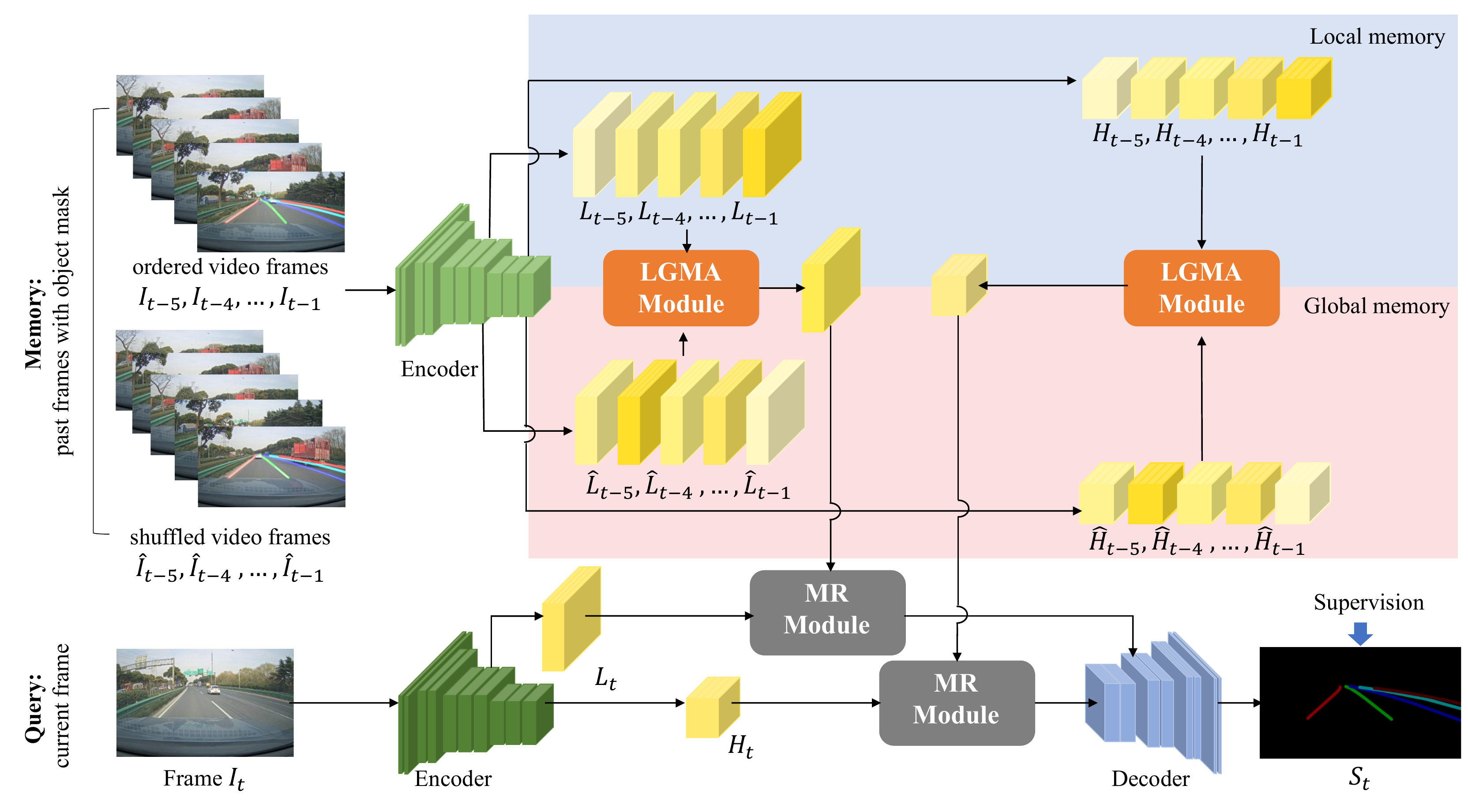}
	\end{center}
	\vspace{-5mm}
	\caption{The schematic illustration of our multi-level memory aggregation network (MMA-Net) for video instance lane detection. ``LGMA'' denotes the local-global memory aggregation module while ``MR'' module is the memory read module.}
	\label{fig:framework}
	\vspace{-1mm}
\end{figure*}

\subsection{Data Collection and Split}

VIL-100 dataset consists of 100 videos, 100 frames per video, in total 10,000 frames. 
The fps rate of all the videos is 10, by down-sampling from 30fps videos.
Among these 100 videos, 97 are collected by monocular forward-facing camera mounted near the rear-view mirror.
%They are collected on various kinds of road in ?? cities by using ?? different vehicles. 
The remaining 3 videos are collected from Internet and they are taken in hazy weather, which increases the complexity and reality of the collected dataset.
We consider 10 typical scenarios in data collection: normal, crowded, curved road, damaged road, shadows, road markings, dazzle light, haze, night, crossroad. Other than the normal one, the latter nine of them usually bring more challenges to lane detection.
We split the dataset to training set and test set according to the ratio of 8:2, and all 10 scenarios are presented in both training and test sets. This can facilitate the consistent use of and fair comparison of different methods on our dataset.

% with different vehicles in multiple cities. In addition, to further obtain richer scenes, we further collect haze scene from the Internet. We manually trim 26 videos into 100 shot clips to make sure diverse scenarios.
% We collect 1 type of simple scenario with clear lanes, and ohter 10 types of challenging scenarios depends on the weather, traffic conditions, and times of the day. To ensure significant and continuous variation in each scenario, we adjust the frame rate from the original 30fps to 10fps. Finally, we obtain our video instance lane dataset (VIL), which includes 100 clips with 100 frames per video.

\subsection{Annotation}

For each video, we place a sequence of points positioned along the center line of each lanes in each frame and store them in json-format files. Points along each lane are stored in one group, which provides the instance-level annotation in our work. We then fit each group of points into a curve by third-order polynomials, and expand it into a lane-region with a certain width. Empirically, on 1,920 $\times$ 1,080 frames, we select the width to be 30 pixels. For lower-resolution frames, the width is reduced proportionally. We further annotate each lane as one of the 10 line-types, i.e., single white solid, single white dotted, single yellow solid, single yellow dotted, double white solid, double yellow solid, double yellow dotted, double white solid dotted, double white dotted solid, double solid white and yellow. In each frame, we also assign an number label to reflect its relative position to the ego vehicle, i.e., an even label $2i$ indicates the $i$-th lane to the right of vehicle while an odd label $2i-1$ indicates the $i$-th lane to the left of vehicle. In VIL-100, we set $i=1,2,3,4$ that enables us to annotate as many as eight lanes in a frame.

% . Also, considering the practicality of lane detection, on each frame, every lane is assigned an identity based on its position relative to the ego vehicle. To adapt to more lane line detection methods, we further fit the discretized points into curves by third-order polynomials, and expand them into regions with a certain width. Empirically, we mark lines with widths equal to 30 pixel for the clips with a resolution of 1920 $\times$ 1080, the discretized points in the clips are scaled up to expand the responding lane region.  
 
\subsection{Dataset Features and Statistics}

% \textbf{Distribution of scenarios.} 

While we consider 10 scenarios in data collection, multiple scenarios may co-occur in the same video, or even in the same frame. Actually 17\% of the video frames contain multiple scenarios, and Figure~\ref{fig:scenarios}~(a) shows the frame-level frequency of such co-occurrence of the 10 scenarios in VIL-100. Meanwhile, one scenario may occur only in part of the video. For example, in video, the scenario may change from `normal' to `crossroad', and then get back to `normal' again in the frames corresponding to `crossroad', there should be no lane detected.
Figure~\ref{fig:scenarios}~(b) shows the total number of frames for each scenario -- a frame with co-occurred scenarios is counted for all present scenarios.

% Inspired by dataset Culane~\cite{SCNN2018}, we collect 11 types of common scenarios on multi-weather, multi-traffic conditions, and different times of the day. Different from the exsiting dataset, we collect a continuous clip with dynamic interference for lane detection. For example, when the vehicle is passing a crossroad, the lanes shouldn't be detected, once passed the crossroad, the lanes should be detected again. The 11 scenarios include one type of simple scene with clear lanes and 10 types of challenging scenes, and the 10 challenging  scenes often exist simultaneously as shown in Figure~\ref{fig:scenarios} (a).  The complex challenging scenes accounted for 17$\%$ of the 100 clips in VIL-100. For a property in a scenario, we count both the separate appeared and the complex appeared. Figure~\ref{fig:scenarios} (b) show the numbers of each type of scenario.

%\textbf{Statistics of lane types.} 

% As mentioned above, we annotate lane instances with the attributes of linetype and relative location to the ego vehicle. We can also count the total number of lanes in a frame based on our annotations. 
As shown in Table~\ref{table:datasets_com}, our VIL-100 contains 10 line-types and provides 6 annotated lanes at most in video frames. %shows the statistics of line-types and the number of annotated lanes in our VIL and existing datasets. 
%Specifically, Figure~\ref{fig:lanes}(a) shows the number of video frames for 10 line-types,
Specifically, Figure~\ref{fig:lanes}(a) shows the number of annotated lanes for 10 line-types,while Figure~\ref{fig:lanes}(b) presents the number of video frames with different annotated lanes, showing that 3,371 video frames have 5 annotated lanes and 13 frames have 6 annotated lanes in our VIL-100.
%the number of As shown in  Figure~\ref{fig:lanes}, we can see that our VIL contains multiple line-types, and some line-types occur more often than the others, as shown in our lane detection dataset. As shown in Figure~\ref{fig:lanes}~(b), We can also see that in VIL-100, 3,371 frames have 5 annotated lanes and 13 frames have 6 annotated lanes.
%while in other public dataset, the number of lanes annotated in each frame is no more than 4.

% As shown in Table~\ref{table:datasets_com}, different from the existing data sets, VIL simultaneously includes multi-linetype, multi-numbers and relative location in the attributes of lanes. Specifically, on the left and right sides of vanishing point, i.e., the intersection of lanes, we define 4 types of semantic identity from near to far,  in total of 8 types of relative location. Due to perspective, the lanes at the far left or right of the vanishing point are too close to be identified, so we annote up to 6 lanes in an image, and the statistics of lane number can be seen in Figure~\ref{fig:lanes}~(a). Also, as shown in Figure~\ref{fig:lanes}~(b) we annote 10 types of common linetype. 
\section{Proposed Method} \label{sec:method}
%%%%%%%%%%%%%%%%%%%%%%%%%%%%%%%%%%%%%%%%%%%%%%%%%%%%%%%%%%%%%%%%%%%%%

%------------------------------------------------------------------------

%%%%%%%%%%%%%%%%%%%%%%%%%%%%%%%%%%%%%%%%%%%
\begin{figure*}
	\begin{center}
		\includegraphics[width=0.90\linewidth]{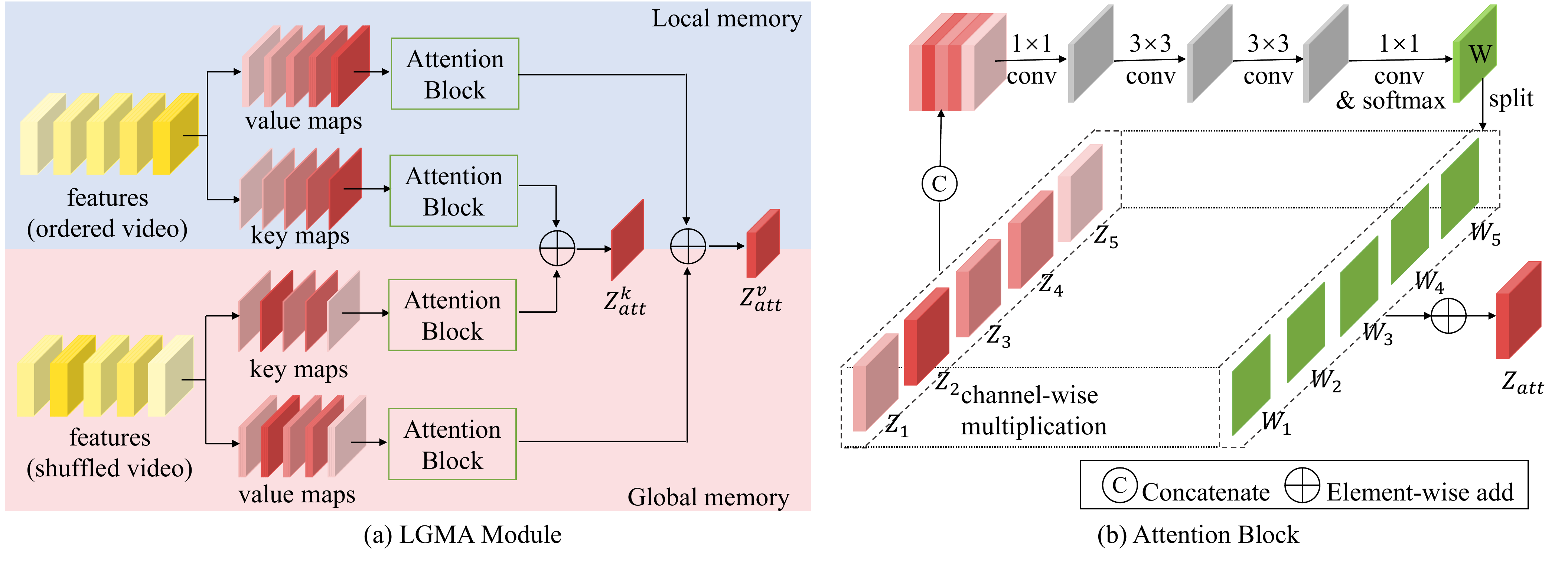}
	\end{center}
    \vspace{-5mm}
	\caption{Schematic illustration of (a) our local and global memory aggregation (LGMA) module, and (b) our attention block. These input five features of LGMA module can be low-level features or high-level features; see Figure~\ref{fig:framework}. And these input features $\{Z_1, Z_2, \ldots, Z_5 \}$ of our attention block can be these five key maps or five value maps of our LGMA module. }
	\vspace{-3mm}
	\label{fig:LGAM}
\end{figure*}

%This section first overviews our MLGM-Net in Section~\ref{sec:MLGMnet}. Section~\ref{sec:attention} shows the details of local and global memory (GLM) module. Section~\ref{sec:implement} provides implementation details of our network.

%\subsection{Our network}
%\label{sec:MLGMnet}

%To fully exploit the temporal consistency for video lane detection, inspired by a semi-supervised video object segmentation~\cite{o6}, we leveraging memory module to store the past relevant information.
%Based on this, to avoid the limitation of local memory, we further aggregate global and local information in the memory module.

Figure~\ref{fig:framework} shows the schematic illustration of our multi-level memory aggregation network (MMA-Net) for video instance lane detection.
Our motivation is to learn memory-based features to enhance low-level and high-level features of each target video frame for video instance lane detection.
The memory features are obtained by integrating a local attentive memory information from the input video and a global attentive memory information from a shuffled video.

Our MMA-Net starts by randomly shuffling an ordered video index sequence $\{1,...,T\}$ of the input video ($T$ frames) to obtain a shuffled index sequence, which is then utilized to generating a shuffled video by taking all corresponding video frames of the input video based on the shuffled video index sequence.
To detect lane regions of a target video frame (i.e., $I_{t}$ of Figure~\ref{fig:framework}), we then take five past frames ($\{I_{t-5},I_{t-4},...,I_{t-1}\}$) of the original video and five past frames ($\{\widehat{I}_{t-5},\widehat{I}_{t-4},..., \widehat{I}_{t-1}\}$) of the shuffled video as the inputs.
Then, we pass each video frame to a CNN encoder consisting of four CNN layers to obtain a high-level feature map ($H$) and a low-level feature map ($L$).
By doing so, we can construct a local memory (denoted as $\mathcal{M}_l$) by storing five low-level features and five high-level features from $\{I_{t-5},I_{t-4},...,I_{t-1}\}$, and form a global memory  (denoted as $\mathcal{M}_g$)  to contain five low-level features and five high-level features from $\{\widehat{I}_{t-5},\widehat{I}_{t-4},..., \widehat{I}_{t-1}\}$.
After that, we develop a local-global memory aggregation (LGMA) module to integrate all low-level features at $\mathcal{M}_l$ and $\mathcal{M}_g$, and another LGMA module to fuse all high-level features at $\mathcal{M}_l$ and $\mathcal{M}_g$. We use $L_{ma}$ and $H_{ma}$ to denote output features of two LGMA modules.
Then, we pass $L_{ma}$ and the low-level features $L_t$ of the target video frame $I_t$ to a memory read (MR) module for enhancing $L_t$ by computing their non-local similarities.
We also refine the high-level features $H_t$ of the target video frame $I_t$ by passing it and $H_{ma}$ into another MR module.
Finally, we follow other memory networks~\cite{o6} to adopt a U-Net decoder to progressively fuse features at different CNN layers and predict a video instance lane detection map for the target video frame $I_t$.
%by using a $1$$\times$$1$ convolutional layer.

\subsection{Local and Global Memory Aggregation Module}
\label{sec:LGAM}

%\vspace{-3mm}

%By providing more video frames in an external memory, memory networks~\cite{o6,chen2020memory,lu2020video,seong2020kernelized,wu2020memory} have achieved superior in many video detection/segmentation tasks.
%%
%Although achieving superior performance in many video detection/segmentation tasks, 
Existing memory networks~\cite{o6,chen2020memory,lu2020video,seong2020kernelized,wu2020memory} utilized a regular sampling on every $N$ frames to include close and distant frames, but all sampled frames are ordered, and extracted features may depend so much on temporal information.
In contrast, we devise a local-global memory aggregation (LGMA) module to utilize five frames from a shuffled video in global memory to remove the temporal order and enhance the global semantic information for detecting lanes.
%%
%aggregate local attentive memory features and global attentive memory features for instance-level video lane detection.
%%
%Compared to the original local memory strategy~\cite{o6}, our LGMA aggregates global semantic information from a shuffled video and local localization information from the input video to segment lane regions of all frames of the input video. \textcolor{red}{Specifically, \cite{o6} utilized a regular sampling on every $N$ frames to include close and distant frames, but all sampled frames are ordered, and extracted features may depend so much on temporal information. In contrast, we utilizes five shuffled frames in global memory to remove the temporal order and enhance the global semantic information for detecting lanes.}
%%
More importantly, due to varied contents of different video frames, memory features from different frames should have varied contributions for helping the current video frame to identify the background objects.
Hence, we leverage an attention mechanism to learn attention maps to automatically assign different weights on both local memory features and global memory features.
%%

% $\{L_{t-5},L_{t-4},...,L_{t-1}\}$ 
% $\{\widehat{L}_{t-5},\widehat{L}_{t-4},..., \widehat{L}_{t-1}\}$

%%%%%%%%%%%%%%%%%%%%%%%%%%%%%%%%%%%%%%%%%%%
Figure~\ref{fig:LGAM}~(a) shows the schematic illustration of the developed LGMA module, which takes five features from the input video and five features from the shuffled video.
We first follow the original memory network~\cite{o6} to extract a pair of key and value maps by applying two $3$$\times$$3$ convolutional layers on each input feature map.
Then, we pass key maps of the local memory to an attention block for a  weighted average on them, fuse key maps of the global memory by passing them into another block, add these features obtained from two attention block to produce a output key map (denoted as $\mathbf{Z}_{att}^{\rm{k}}$) of the LGMA module.
Meanwhile, we generate of the output value map (denoted as $\mathbf{Z}_{att}^{\rm{v}}$) of our LGMA module by adding these generated features of two attention blocks, which aim to aggregate the value maps of the local memory and the global memory, respectively. 
%Meanwhile, another two attention block and key maps the global memory together to an attention block to weighted average them, and another attention block is devised to aggregate value maps of the local memory and the global memory.
%%
Mathematically, the output key map $\mathbf{Z}_{att}^{\rm{k}}$ and the output value map $\mathbf{Z}_{att}^{\rm{v}}$ of our LGMA module are computed as:
\vspace{-1mm}
\begin{equation}
\label{eq:attentive_features}
\begin{aligned}
\mathbf{Z_{att}^{\rm{k}}}= f_{att}(\mathbf{k_1^L}, \mathbf{k_2^L}, \ldots, \mathbf{k_5^L}) + f_{att}(\mathbf{k_1^G}, \mathbf{k_2^G}, \ldots, \mathbf{k_5^G}) \ , \\
\mathbf{Z_{att}^{\rm{v}}} = f_{att}(\mathbf{v_1^L}, \mathbf{v_2^L}, \ldots, \mathbf{v_5^L}) + f_{att}(\mathbf{v_1^G}, \mathbf{v_2^G}, \ldots, \mathbf{v_5^G}) \ ,
\end{aligned}
\end{equation}
\vspace{-1mm}
where $f_{att}(\cdot)$ denotes an attention block to attentively integrate memory features.
$\{ \mathbf{k_1^L}, \mathbf{k_2^L}, \ldots, \mathbf{k_5^L} \}$ and $\{ \mathbf{v_1^L}, \mathbf{v_2^L}, \ldots, \mathbf{v_5^L} \}$ denote key maps and value maps of five input features of the local memory.
$\{ \mathbf{k_1^G}, \mathbf{k_2^G}, \ldots, \mathbf{k_5^G} \}$ and $\{ \mathbf{v_1^G}, \mathbf{v_2^G}, \ldots, \mathbf{v_5^G} \}$ are key maps and value maps of five input features of the global memory.
As shown in our framework of Figure~\ref{fig:framework}, we pass these low-level features of both the local memory and the global memory into a LGMA module to aggregate them for generating a pair of key map and value map (denoted as $L(\mathbf{Z}_{att}^{\rm{k}}$) and $L(\mathbf{Z}_{att}^{\rm{v}}$)).
Also, another LGMA module is devised to aggregate the high-level features of both the local memory and the global memory, and these two output key and value maps are denoted as $G(\mathbf{Z}_{att}^{\rm{k}})$ and $G(\mathbf{Z}_{att}^{\rm{v}})$; see Figure~\ref{fig:framework}.

%Moreover, as shown in Fig.~\ref{fig:framework}, another LGMA module is developed to attentively fuse five high-level features of the input video and five high-level features of the shuffled video to generate output key and value maps (denoted as $\mathbf{Z}_{att}^{\rm{k}}$ and $\mathbf{Z}_{att}^{\rm{v}}$). 

%%%%%%%%%%%%%%%%%%%%%%%%%%%%%%%%%%%%%%%%%%%
\vspace{2mm}
\noindent  \textbf{Attention block.} \ Figure~\ref{fig:LGAM}~(b) shows the developed attention block to attentively integrate input five feature maps $\{Z_1, Z_2, \ldots, Z_5 \}$, which can be the five key maps or five value maps features of Figure~\ref{fig:LGAM}~(a).
Specifically, we first concatenate five input maps and then utilize a $1$$\times$$1$ convolutional layer, two successive $3$$\times$$3$ convolutional layer, a $1$$\times$$1$ convolutional layer, and a Softmax function on the concatenated feature map to produce an attention map $W$ with five feature channels.
Then, we multiply each channel of $W$ with these input five maps, and then we add these multiplication results together to produce an output map ($Z_{att}$) of the attention block.
Hence, $Z_{att}$ is computed as
\begin{equation}
\label{eq:attention}
Z_{att} = \sum_{i=1}^{5} (\mathbf{W}_i \bigotimes \mathbf{Z_i}) \ ,
\end{equation}
where $\{Z_1, Z_2, \ldots, Z_5 \}$ denotes all five input maps of our attention block, and they can be five key maps or five value maps of LGMA module; see Figure~\ref{fig:LGAM}~(a).
$W_i$ is the $i$-th channel of the attention map W.
$\bigotimes$ is the multiplication of $W_i$ and $Z_i$.

%%%%%%%%%%%%%%%%%%%%%%%%%%%%%%%%%%%%%%%%%%%
\subsection{Implementation Details}
\label{sec:implement}

%To extract features to construct the local and global memory,

%%%%%%%%%%%%%%%%%%%%%%%%%%%%%%%%%%%%%%%%%%%
%\vspace{2mm}
\noindent  \textbf{Memory read module.} \  Following the original memory network~\cite{o6}, we also develop a memory read (MR) module to retrieve the relevant memory information (i.e, the key and value map of our LGMA module; see Figure~\ref{fig:LGAM}~(a)) for the query frame (i.e., the target video frame $I_t$ of Figure~\ref{fig:framework}).
Specifically, we first apply two $3$$\times$$3$ convolutional layers on  features of the query frame $I_t$ to obtain a pair of key map and value map,
Then, the MR module first obtains an non-local affinity matrix by computing similarities between all pixels of the output key map of our LGMA module and the key map of $I_t$.
After that, we multiple the affinity matrix with the output value map of our LGMA module, and
then concatenate the multiplication result with the value map of $I_t$ to produce the output features of the MR module.

%We employ a spatial attention (SA) module to retrieve the relevant memory information for the query frame. Specifically, similarities between key maps of the query and the memory frames are first computed to determine where to retrieve relevant memory values from, the affinity matrix $\textbf{A}$ is performed in a non-local manner by comparing every space location in the memory key map $\mathbf{k}^{\rm{M}}$ with every spatial location in the query key map $\mathbf{k}^{\rm{Q}}$, as $\mathbf{A}=\rm{softmax}(\frac{\mathbf{k}^Q(\mathbf{k}^M)^{\rm{T}}}{\sqrt{C/8}})$. Then, the memory value map $\mathbf{v}^{\rm{M}}$ is retrieved by a weighted summation with the similarity matrix and it is concatenated with the query value $\mathbf{v}^{\rm{Q}}$.

%%%%%%%%%%%%%%%%%%%%%%%%%%%%%%%%%%%%%%%%%%%
\vspace{2mm}
\noindent  \textbf{Decoder.} \  As shown in Figure~\ref{fig:framework}, our network employs two memory read (MR) modules to read the corresponding attentive memory features to enhance features at the 3rd CNN layer and the 4-th CNN layer.
After that, the decoder of our network takes the output features of two MR modules to predict the instance-level lane detection result of the target video frame $I_t$.
To do so, we first compress the output features of two MR modules to have $256$ channels by a convolutional layer and a residual block.
Then, three refinement blocks (see~\cite{o6} for the details of the refinement block) are employed to gradually fuse two compressed feature maps and these two encoder features at the first two CNN layer, and each refinement block upscales the spatial resolution by a factor of two.
Finally, we follow~\cite{o6} to produce the video instance-level lane detection result from the output features of the last refinement block.
%by utilizing a Softmax function on the output features to produce a mask probability map (8$\times$H$\times$W) for all lanes.
%Specifically, we independently compute mask probability map for each lane instance, then, similar to~\cite{softagg}, we merge the predicted maps using a soft aggregation operation.

%%%
%The first refinement takes the two compressed features as the inputs, and the second refinement take the output of the first refinement
%Then, a refinement module is utilized to take two compressed features as the input, and the obtained features are
%
%number of refinement modules are employed to gradually upscale the
%compressed feature maps by a factor of two at a time.
%Each refinement module takes both the output of the
%previous stage and a feature map from the query encoder
%at the corresponding scale through skip-connections. The
%output of the last refinement block is used to reconstruct the
%object mask through the final convolutional layer followed
%by a softmax operation. Every convolutional layer in the
%decoder uses 3×3 filters, producing 256-channel output except
%for the last one that produces 2-channel output. The
%decoder estimates the mask in 1/4 scale of the input image.

\vspace{2mm}
\noindent  \textbf{Our training procedure.} \ Like~\cite{o6}, we first train the feature extraction backbone (i.e., encoder of Figure~\ref{fig:framework}) of our network to extract features for each video frame. 
Specifically, we take the past two frames (only from the input video) of the current video frame (query frame) to construct a local memory, and then employ a memory read (MR) module to read the local memory feature for producing a instance-level lane detection result of the query frame.
After that, we take five past frames from the input video and five past frames of a shuffled video of the current video frame (query frame), and the encoder trained in the first training stage to obtain their feature maps, and then follow the network pipeline of Figure~\ref{fig:framework} to predict an instance-level lane detection result of the target video frame to train our network.
In these two training stages, we empirically add a cross entropy loss and a IoU loss to compute the loss of the predicted instance-level lane map and the corresponding ground truth for training.

\vspace{2mm}
\noindent \textbf{Training parameters.} \
We implement our MMA-Net using Pytorch and train our network on a NVIDIA GTX 2080Ti.
In the first training stage, we initialize the feature extraction backbone by using pre-trained ResNet-50~\cite{ResNet50}, and employ Adam optimizer with a learning rate of $10^{-5}$, a momentum value of $0.9$, and a weight decay of $5\times10^{-4}$.
In the second training stage, stochastic gradient descent optimizer is employed to optimize the whole network with the learning rate as $10^{-3}$, a momentum value of $0.9$, a weight decay of $10^{-6}$, and a mini-batch size of $1$.
%Moreover, we add random noise on input images for data argumentation.
The first training stage takes about 14 hours with $100$ epochs while the second training stages takes about 7 hours with $50$ epochs.
%with a mini-batch size of 1

%On VIL, with an NVIDIA 2080Ti, MLGM-Net achieves 12.4fps.

\section{Experiments}
\label{sec:experiments}

%%%%%%%%%%%%%%%%%%%%%%%%%%%%%%%%%%%%%%%%%%%
\begin{table*}[t]
	\caption{Quantitative comparisons of our network and state-of-the-art methods in terms of image-based metrics.}
	\vspace{-3mm}
	\centering
	\begin{tabular}{c|c|ccc|ccc}
		\hline
		\multirow{2}{*}{\textbf{Methods}} & \multirow{2}{*}{\textbf{Year}}
		& \multicolumn{3}{c|}{\textbf{Region}} & \multicolumn{3}{c}{\textbf{Line}}  \\ \cline{3-8}
		& & \textbf{mIoU$ \uparrow$} & \textbf{F1$^{0.5}$ $\uparrow$} & \textbf{F1$^{0.8}$ $\uparrow$} & \textbf{Accuracy $\uparrow$} & \textbf{FP $\downarrow$} & \textbf{FN $\downarrow$}                            \\ \hline
		LaneNet~\cite{LaneNet2018} &2018 & 0.633& 0.721 & 0.222 & 0.858& 0.122 & 0.207 \\
		SCNN~\cite{SCNN2018} &2018 & 0.517& 0.491 & 0.134 & 0.907& 0.128 & 0.110  \\
		ENet-SAD~\cite{r11} &2019 & 0.616& 0.755 & 0.205 & 0.886& 0.170 & 0.152\\
		UFSA~\cite{r14} & 2020 & 0.465& 0.310 & 0.068 & 0.852& 0.115 & 0.215  \\
		LSTR~\cite{r17} & 2021 & 0.573  & 0.703 & 0.131 & 0.884  & 0.163 & 0.148 \\
		\hline
		GAM~\cite{GAM} &2019 & 0.602& 0.703 & 0.316 & 0.855& 0.241 & 0.212 \\
		RVOS~\cite{RVOS} &2019 & 0.294 & 0.519 & 0.182 & 0.909 & 0.610 & 0.119  \\
		STM~\cite{o6}&2019 & 0.597 & 0.756 & 0.327 & 0.902 & 0.228 & 0.129 \\
		AFB-URR~\cite{AFB} &2020 & 0.515 & 0.600 & 0.127 & 0.846 & 0.255 & 0.222 \\
		TVOS~\cite{TVOS} & 2020& 0.157& 0.240 & 0.037 & 0.461& 0.582 & 0.621 \\
		\hline \hline
		MMA-Net (Ours) &2021 & \textbf{0.705} & \textbf{0.839} & \textbf{0.458} & \textbf{0.910} & \textbf{0.111} & \textbf{0.105} \\ \hline
	\end{tabular}
	\label{table:res_compare}
\end{table*}

%Note that Tusimple~\cite{Tusimple2017} and Zou et al.~\cite{zou2019robust} also collects lane videos, but only annotates lane regions of the last frame of each video, or the 13th frame and the last frame.
%%%
%ApolloScape~\cite{ApolloScape2019} and BDD100K~\cite{BDD100K2020} are two binary video lane detection datasets, which different lanes are assigned with same label. 
%only to build a dataset for single-image lane detection.

%%%%%%%%%%%%%%%%%%%%%%%%%%%%%%%%%%%%%%%%%%%
%\vspace{2mm}
\noindent \textbf{Dataset.} \ 
Currently, there is no benchmark dataset dedicated for training video instance lane detection by annotating instance-level lanes of all frames in videos.
%%
%Hence, we collect a dataset (i.e, VIL-100; see Section~\ref{section3:dataset}) with $100$ videos, and each frame of all videos are with instance-level annotations on lane regions; see Figure~\ref{fig:dataset} for examples. 
%%
With our VIL-100, we test the video instance lane detection performance of our network and compared methods. 
%%
%Since the MLGM-Net is based on video-level lane segmentation, common benchmarks are Culane and Tusimple, but they are lack of continuity of the sequence. As a result, we conduct all experiments on our VIL Datesets.

%%%%%%%%%%%%%%%%%%%%%%%%%%%%%%%%%%%%%%%%%%%
\vspace{2mm}
\noindent \textbf{Evaluation metrics.} \ To quantitatively compare different methods, we first employ six widely-used image-level metrics, including three region-based metrics and three line-based metrics.
Three region-based metrics~\cite{SCNN2018,chen2021triple} are mIoU,  F1(IoU\textgreater{}0.5) (denoted as F1$^{0.5}$), and F1(IoU\textgreater{}0.8) (denoted as F1$^{0.8}$), while three line-based metrics are Accuracy, FP, and FN.
Apart from image-level metrics~\cite{Tusimple2017}, we also introduce video-level metrics to consider the temporal stability of the segmentation results for further quantitatively comparing different methods.
Video-level metrics are \textbf{$\mathcal{M_J}$}, \textbf{$\mathcal{O_J}$}, \textbf{$\mathcal{M_F}$}, \textbf{$\mathcal{O_F}$}, and \textbf{$\mathcal{M_T}$}; please refer to~\cite{perazzi2016benchmark} for definitions of these video-level metrics.
In general, a better video instance lane detection method shall have larger mIoU, F1$^{0.5}$, F1$^{0.8}$, and accuracy scores, as well as smaller FP and FN scores.
According to~\cite{perazzi2016benchmark}, a better video instance segmentation method has larger scores for all video-based metrics.

%Since there is no CNN-based method for video lane,

\vspace{2mm}
\noindent \textbf{Comparative methods.}  \
To evaluate the effectiveness of the developed video instance lane detection method, we compare it against $10$ state-of-the-art methods, including LaneNet~\cite{LaneNet2018}, SCNN~\cite{SCNN2018}, ENet-SAD~\cite{r11}, UFSA~\cite{r14}, LSTR~\cite{r17}, GAM~\cite{GAM}, RVOS~\cite{RVOS}, STM~\cite{o6}, AFB-URR~\cite{AFB} and TVOS~\cite{TVOS}.
Among them, LaneNet, SCNN, ENet-SAD, UFSA, and LSTR are image-level lane detection methods, while GAM, RVOS, STM, AFB-URR and TVOS are instance-level video object detection.
Since our work focuses on video instance lane detection, we do not include video binary segmentation methods (e.g., video salient object detection, video shadow detection) for comparisons.
For all comparing methods, we use their public implementations, and re-train these methods on our VIL-100 dataset for a fair comparison.

% aim to not only segment lane pixels, but also different labels should be assigned for multiple lanes.
%Hence, we compare our network with instance-level video object segmentation methods, and 

%%%%%%%%%%%%%%%%%%%%%%%%%%%%%%%%%%%%%
\begin{figure*}
	\begin{center}
		\includegraphics[width=0.82\linewidth]{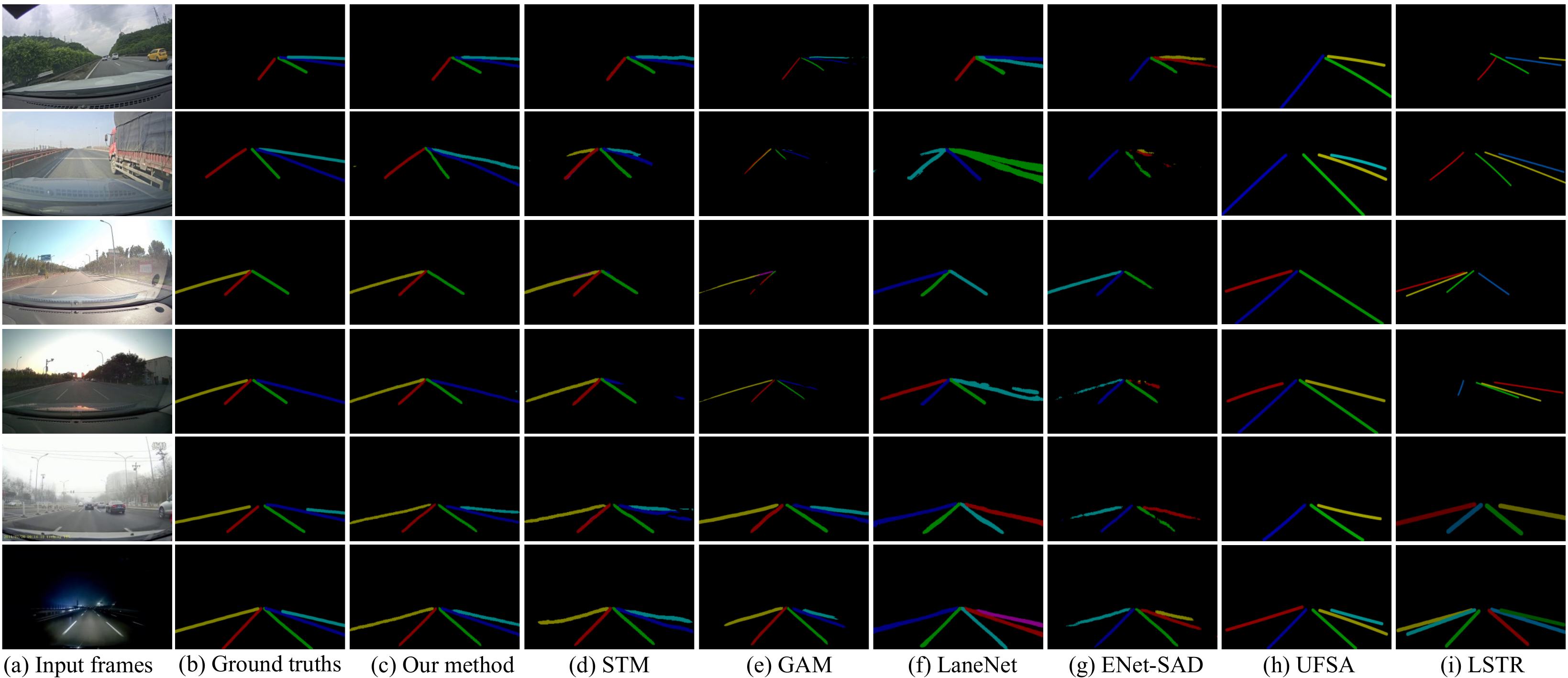}
	\end{center}
    \vspace{-5mm}
	\caption{Visual comparison of video instance lane detection maps produced by our network (3rd column) and state-of-the-art methods (4-th to 9-th columns) against ground truths (2nd column). Please refer to supp. material for more comparisons.}
	\label{fig:visualcomp}
	\vspace{-2mm}
\end{figure*}

%%%%%%%%%%%%%%%%%%%%%%%%%%%%%%%%%%%%% Table 3
\begin{table}[t]
	\centering
	\caption{Quantitative comparisons of our network and state-of-the-art methods in terms of video-based metrics.}
	\vspace{-3mm}
	\resizebox{\linewidth}{!}{
	\begin{tabular}{c|ccccc}
		\hline
		\textbf{Methods} & \textbf{$\mathcal{M_J}$ $\uparrow$} & \textbf{$\mathcal{O_J}$  $\uparrow$} &  \textbf{$\mathcal{M_F}$ $\uparrow$} & \textbf{$\mathcal{O_F}$ $\uparrow$} &  \textbf{$\mathcal{M_T}$ $\uparrow$} \\
		\hline
        GAM~\cite{GAM} & 0.414 & 0.203 & 0.721 & 0.781 & 0.568\\
        RVOS~\cite{RVOS} & 0.251 & 0.251 & 0.251 & 0.251 & 0.251  \\		
		STM~\cite{o6} & 0.656 & 0.626 & 0.743 & 0.763 & 0.656  \\
		AFB-URR~\cite{AFB} & 0.308 & 0.251 & 0.415 & 0.435 & 0.362 \\
		TVOS~\cite{TVOS} & 0.255 & 0.251 & 0.257 & 0.256 & 0.255  \\
		 \hline \hline
		\tabincell{c}{MMA-Net\\(Ours)} & \textbf{0.679} & \textbf{0.735} & \textbf{0.848} & \textbf{0.873} & \textbf{0.764} \\ \hline
	\end{tabular}}
	\vspace{-3mm}
	\label{table:fb_compare}
\end{table}

\subsection{Comparisons with State-of-the-art Methods}

%%%%%%%%%%%%%%%%%%%%%%%%%%%%%%%%%%%%%%%%%%%
%\vspace{2mm}
\noindent \textbf{Quantitative comparisons.} Table~\ref{table:res_compare} reports six image-level quantitative results of our network and all compared methods.
%%
%\textcolor{red}{
Basically, we can observe that lane detection methods have a better performance on line-based metrics, since they often utilize line-related information (e.g., shape and direction) to infer the lines. By contrast, the VOS methods formulate the lane detection task as a region-based segmentation with abjectness constraint and thus perform better on region-based metrics.
%}
%Note that our method obtains the best performances on all line/region-based metrics.
%shows the quantitative evaluation among our method and others published in the last three years, e.g., the start-of-the-art network Lane-Net.
%Apparently, our method outperforms all competitors on all six metrics.
%%As shown in Table~\ref{table:res_compare}, 
%Among all these compared methods,
Specifically, LaneNet has the best mIoU score of $0.633$, STM has the best F1$^{0.5}$ of $0.756$, and the best F1$^{0.8}$ of $0.327$.
Regarding Accuracy, FP, and FN, RVOS has the best Accuracy of $0.909$; UFSA has the best FP of $0.115$; and SCNN has the best FN of $0.110$; see Table~\ref{table:res_compare}.
Compared to these best scores of different metrics, our method has a mIoU improvement of $11.37\%$,
a F1$^{0.5}$ improvement of $10.98\%$, a F1$^{0.8}$ improvement of $40.06\%$, a Accuracy improvement of $0.11\%$,
a FP improvement of $3.48\%$, and a FN improvement of $4.55\%$.
%Notes that even though all other methods have attained quite high result of Accuracy, the value of our method is still 5$\%$ higher than the average, while our FN value also attains the lowest value. At the same time, our results show a 10$\%$ - 20$\%$ improvement in mIoU. Besides, we notice that with the increase of IoU setting range, the gap with F1 between ours and others widens.
%
%In addition, among the methods of lane detection, the most challenging baseline is UFSA, whose FP value is the lowest. But in fact this task has a apriority and dependence on the row-anchors, moreover the size of input images will also cause the fluctuation of the outputs. Notes that although UFSA and LSTR only output points and lines respectively, both mIoU and F1 have been expanded to 30pixels in our calculations, so the comparison is fair. As for STM, though it is competitive in the video segmentation methods, it needs to input the GT of the first mask in testing, which is not practical for the real circumstance.

Moreover, Table~\ref{table:fb_compare} summaries video-based metric scores of our network and compared methods.
Among results of compared video-based methods, we can find that GAM has the largest $\mathcal{O_F}$ score (i.e., 0.781), while
STM has the best performance of other four video-based metrics.
These corresponding best four values of STM are $\mathcal{M_J}$ of $0.656$,  $\mathcal{O_J}$ of $0.626$,  $\mathcal{M_F}$ of $0.743$,  and  $\mathcal{M_T}$ of $0.656$.
More importantly, our method achieves a further improvement for all five video-based metrics, showing that our method can more accurately segment lanes of different videos.
To be specific, our method improves $\mathcal{M_J}$ from $0.656$ to 0.679; $\mathcal{O_J}$ from $0.626$ to 0.735; $\mathcal{M_F}$ from $0.743$ to $0.848$; $\mathcal{O_F}$ from 0.781 to $0.873$; and $\mathcal{M_T}$ from $0.656$ to $0.764$.

%shows that our video segmentation network has stronger advantages under different video segmentation networks. Obviously, the value of J, F, and FM go far beyond the other methods. Among these, our method is the same as AFB-URR and RVOS which are all unsupervised. What is interesting is that compared with the two methods, our FM is largely improved by 2.3 times on average, while J and F also is improved by nearly three times in value. It is observed that STM still has definite advantages and competitiveness among these methods, but its disadvantages cannot be ignored for the actual situation, and likewise, TVOS and GAM also have the same disadvantages.

%%%%%%%%%%%%%%%%%%%%%%%%%%%%%%%%%%%%%%%%%%%
\vspace{2mm}
\noindent \textbf{Visual comparisons.} \ Figure~\ref{fig:visualcomp} visually compares video instance lane detection maps produced by our network and compared methods.
Apparently, compared methods neglect some lane regions or wrongly recognized parts of road regions as target lane regions, as shown in 4-th to 9-th columns of Figure~\ref{fig:visualcomp}.
Moreover, instance labels of different lanes are also mistakenly detected in video instance lane detection results of compared methods.
On the contrary, our method can more accurately detect lane regions and has correct instance labels for all lanes, and our results are more consistent with the ground truths of Figure~\ref{fig:visualcomp}~(b).
In addition, for these challenging cases (i.e, traffic scenes at night or haze weather conditions) at the last two rows, our method also predicts more accurate lane detection maps than competitors, showing the robustness and effectiveness of our video instance lane detection method.

%\textcolor{red}{Note, different VOS methods take diverse frames as their input. The existing VOS methods usually sampled 3/5/7 (less than 10) neighboring frames as the input due to the limitation of GPU memory, while memory-based methods (e.g., \cite{o6}) required 20 frames in the memory to process a video with 100 frames. }

\begin{table}[t]
\centering
\caption{Quantitative results of our network with different sampling numbers.}
\vspace{-3mm}
\resizebox{0.50\textwidth}{!}{%
\vspace{-28pt}
\begin{tabular}{c|ccccc}
\hline
 & \textbf{mIoU $\uparrow$}           & \textbf{\tabincell{c}{F1$^{0.5}$ $\uparrow$}}             & \textbf{Accuracy $\uparrow$}            & \textbf{FP $\downarrow$}             & \textbf{FN $\downarrow$}             \\
\hline
3 frames  & 0.678          & 0.816          & 0.904          & 0.125          & 0.116          \\

5 frames (ours)  & \textbf{0.715} & 0.838          & 0.907          & \textbf{0.106} & 0.111          \\

7 frames  & 0.705          & \textbf{0.839} & \textbf{0.910} & 0.111          & \textbf{0.105}
\\
\hline
\end{tabular}
}
\vspace{-13pt}
\label{table:sam_number}
\end{table}

\noindent \textbf{Sampling number.}
%Note that different VOS methods take diverse frames as their input. The 
Existing VOS methods usually sampled $3/5/7$ (less than 10) neighboring frames as the input due to the limitation of GPU memory, while memory-based methods (e.g., [30]) required 20 frames in the memory to process a video with 100 frames. 
Moreover, we also provide an experiment of our network with $3/5/7$ frames in the Table~\ref{table:sam_number}, where our method with $5$ frames outperforms that with $3$ frames significantly, and is comparable with the one employing $7$ frames.
By balancing the GPU memory and computation consuming, we empirically use $5$ frames in our method.

%{\color{red} TODO...}
%It is clear that our results are robust in challenging scenarios and harsh environments, such as occlusion, darkness, fog, and shadow, with almost no missed or error detected cases, not like LSTR and USFA. Furthermore, because our method combines the information of historical frames, and contains the context information of the current frame as well, it can not only ensure the accuracy of detection, but also avoid inter-frame jump and instability in single-frame lane line detection. As a result, it suggests the significance of video lane segmentation in the lane detection task.

%% Table4
\begin{table}[t]
	\centering
	\caption{Quantitative results of our network and constructed baseline networks of ablation study in terms of image-level and video-level metrics.}
	\vspace{-3mm}
	\begin{tabular}{c|ccccc}
		\hline
		\textbf{Measure} & \textbf{Basic} & \textbf{+LM} & \textbf{+GM} & \textbf{+LGM} & \textbf{Ours}\\ \hline
		\textbf{mIoU $\uparrow$} & 0.638 & 0.688 & 0.670 & 0.693 & \textbf{0.705}\\
		\textbf{\tabincell{c}{F1$^{0.5}$ $\uparrow$}} & 0.758 & 0.790 & 0.796 & 0.822 & \textbf{0.839}\\
		\textbf{\tabincell{c}{F1$^{0.8}$ $\uparrow$}} & 0.402 & 0.425 & 0.423 & 0.450 & \textbf{0.458}\\\hline\hline
		\textbf{Accuracy $\uparrow$} & 0.857 & 0.862 & 0.887 & 0.897 & \textbf{0.910}\\
		\textbf{FP $\downarrow$} & 0.195 & 0.128 & 0.150 & 0.122 & \textbf{0.111}\\
		\textbf{FN $\downarrow$} & 0.173 & 0.163 & 0.136 & 0.124 & \textbf{0.105}\\ \hline\hline
		\textbf{$\mathcal{M_T}$ }$\uparrow$ & 0.708 & 0.706  & 0.721 & 0.760 & \textbf{0.764} \\ \hline
		\textbf{$\mathcal{M_J}$  }$\uparrow$ & 0.627 & 0.632 & 0.640 & 0.678 & \textbf{0.679}\\
		\textbf{$\mathcal{O_J}$  }$\uparrow$ & 0.664 & 0.676 & 0.679 & 0.729 & \textbf{0.735}\\ \hline
		\textbf{$\mathcal{M_F}$  }$\uparrow$ & 0.789 & 0.781 & 0.802 & 0.842 & \textbf{0.848}\\
		\textbf{$\mathcal{O_F}$  }$\uparrow$ & 0.811 & 0.795 & 0.826 & 0.865 & \textbf{0.873}\\ \hline
		%\textbf{FPS $\uparrow$} & 15.2 & 16.0 & 16.4 & 15.1 & 12.4\\ \hline
	\end{tabular}
	\vspace{-3mm}
	\label{table:abl_compare}
\end{table}

\subsection{Ablation Study}

%In this part, we conduct ablation study experiments to evaluate the effectiveness of major components (i.e., local and global memory aggregation module, and multi-level integration mechanism) of our method.

\vspace{0.5mm}
\noindent \textbf{Basic network design.} \
Here, we construct four baseline networks. 
The first one (denoted as ``Basic'') is to remove the attention mechanism from the local memory, the whole global memory, and the multi-level aggregation mechanism from our network. It means that ``Basic'' is equal to the original STM but removes the mask initialization of the first video frame.
The second one (``+LM'') is to add the attention mechanism to the local memory of ``Basic'' to weighted average local memory features,
while the third one (``+GM'') is to add the attentive global memory to ``Basic''.
The last baseline network (``+LGM'') is to fuse the global memory and the local memory together into ``Basic''. It means that we remove the multi-level integration mechanism from our network to construct ``LGM''.
Table~\ref{table:abl_compare} reports image-based and video-based metric results of our method and compared baseline networks.

\vspace{0.5mm}
\noindent \textbf{Effectiveness of the attention mechanism in memory.} \ As shown in Table~\ref{table:abl_compare}, ``+LM'' outperforms ``Basic'' on all image-level and video-level metrics.
It indicates that leveraging the attention mechanism to assign different weights for all memory features, which enables the local memory to extract more discriminative memory features for the query feature refinement, thereby resulting in a superior improving video instance lane detection performance.

\vspace{0.5mm}
\noindent \textbf{Effectiveness of the global memory.} \ ``GM'' has a better performance of image-based metrics and video-based metrics than ``Basic'', demonstrating that the global memory has a contribution to the superior performance of our method.
Moreover, ``LGM'' has superior metrics on all metrics over ``LM'' and ``GM''. It shows that aggregating the local memory and the global memory can further enhance the query features of the target video frame and thus incurs a superior video instance lane detection performance.

\vspace{0.5mm}
\noindent \textbf{Effectiveness of our multi-level mechanism.} \ 
As shown in the last two columns of Table~\ref{table:abl_compare}, our method has larger mIoU, F1$^{0.5}$, F1$^{0.8}$, and Accuracy, smaller FP and FN scores, as well as larger video-based metric ($\mathcal{M_J}$, $\mathcal{O_J}$, $\mathcal{M_F}$, $\mathcal{O_F}$, and $\mathcal{M_T}$) scores than ``+LGM''.
It indicates that applying our LGMA modules on low-level and high-level features of the target video frame enable our network to better detect video instance lanes.

%%
%In our experiment, we consider the STM in which we add the initialization to remove the need of first mask as our basic network(named Basic). Additionally, LM is to add the local memory module into    basic, similarly. GM is to add the global memory module into basic, LGM is add the local and global aggregation memory into basic. Finally, ours adds the multi-level mechanism on the basis of LGM. All implementation of ours follows that shown in Table ~\ref{table:abl_compare}.
%Fig. 8 also visually shows the the difference between adding global and local modules separately and aggregating them. We have the following observations: (1) Most indicators of LM and GM are higher than Basic, which means both local and globel information can be used can provide more effective information for lane detection. (2) the metric of LM are all higher than GM, which can clearly see that... (3) LGM have superior performance of all metrics than to LM and GM, that signifies that the combining of globle and local information have better guidance and promotion effects than seperate.
%%
%it is obvious that the metric of multi-level spatial mechanism are overlap the single scale feature. From results we can deduce that, the fusion of high-level and low-level features can better capture the lane shape and accurare curve.

\section{Conclusion}
\label{sec:conclusion}

%%%%%
%%
To facilitate the research on video instance lane detection, we collected a new VIL-100 video dataset with high-quality instance-level lane annotations over all the frames. VIL-100 consists of $100$ videos with $10,000$ frames covering various line-types and traffic scenes.
%, with instance-level lane annotations.
%%
Meanwhile, we developed a video instance lane detection network MMA-Net by aggregating local attentive memory information of the input video and global attentive memory information of a shuffled video as a new baseline on VIL-100.
Experimental results demonstrated that MMA-Net outperforms state-of-the-art methods by a large margin.
%%
%\textcolor{red}{
We agree that more diverse
scenes/viewpoints could enhance the dataset, and we definitely
continue to collect more data in our future work.
%}
%%
%To the best of our knowledge, this work is the first annotated dataset for video instance lane detection, and our VIL-100 facilitates further research in video instance-level lane detection. 

\section*{Acknowledgments}
The work is supported by the National Natural Science Foundation of China (Project No.  62072334, 61902275, 61672376, U1803264).

%\clearpage

{\small
\bibliographystyle{ieee_fullname}
\bibliography{paper}
}

\end{document}